\documentclass[11pt]{article}

\usepackage[final]{acl}
\usepackage{cuted}
\usepackage{times}
\usepackage{latexsym}
\usepackage{booktabs}
\usepackage[T1]{fontenc}
\usepackage{makecell}

\usepackage{tabularx}

\usepackage[utf8]{inputenc}

\usepackage{microtype}

\usepackage{inconsolata}
\usepackage{cuted}
\usepackage{graphicx}

\usepackage{amsmath}
\usepackage{amssymb}
\usepackage{array}
\usepackage{xcolor}      
\usepackage{colortbl}    
\usepackage{multirow}    
\usepackage{booktabs}
\usepackage{pifont}
\usepackage{cleveref}
\usepackage{tcolorbox}
\usepackage{makecell}
\usepackage{svg}
\usepackage{siunitx}
\usepackage{arydshln}
\usepackage{threeparttable}
\usepackage{wrapfig}
\usepackage[table]{xcolor}
\usepackage{hyperref}
\usepackage{url}
\usepackage{float}
\usepackage{placeins}
\usepackage{fontawesome5}
\usepackage{enumitem}
\usepackage[ruled,vlined,linesnumbered]{algorithm2e}
\usepackage{tikz}
\usepackage{stfloats}
\title{Code-on-Graph: Iterative Programmatic Reasoning via Large Language Models on Knowledge Graphs}

\author{
Weiwei Ding$^{1,2\dagger}$, Zixuan Li$^{1*}$, Long Bai$^{1}$, Zhuo Chen$^{1}$, Kun Su$^{2,3}$, Fei Wang$^{1}$\\
{\bf Xiaolong Jin$^{1*}$, Jin Zhang$^{1}$, Jiafeng Guo$^{1}$} and {\bf Xueqi Cheng$^{1}$} \\
$^{1}$Key Laboratory of AI Safety, Institute of Computing Technology, Chinese Academy of Sciences \\
$^{2}$Shandong University \\
$^{3}$Shandong University-Weihai Research Institute of Industrial Technology \\
}

\newcommand \footnoteONLYtext[1]
{
	\let \mybackup \thefootnote
	\let \thefootnote \relax
	\footnotetext{#1}
	\let \thefootnote \mybackup
	\let \mybackup \imareallyundefinedcommand
}

\begin{document}
\graphicspath{{./}}
\maketitle
\footnoteONLYtext{ 
{\noindent$^{*}$}Corresponding authors.\\ {$^{\dagger}$}This work was completed during Weiwei Ding's internship at the Institute of Computing Technology, Chinese Academy of Sciences. }

\begin{abstract}

Knowledge Graphs (KGs) are widely used to mitigate the limitations of Large Language Models (LLMs), such as outdated knowledge and hallucinations. Existing LLM–KG integration frameworks typically rely on predefined operators to retrieve factual knowledge from KGs and inject it into prompts for answer generation.
This paradigm faces two critical bottlenecks: 
1) \textbf{Inflexibility}: 
The predefined operators are limited in scope and thus lack sufficient compositional expressiveness to fully capture the complex semantics required by KG questions.
2) \textbf{Unscalability}: Direct injection of factual knowledge into prompts limits scalability in handling large-scale factual knowledge.
To address these two bottlenecks, we propose Code-on-Graph (CoG), a programmatic reasoning framework for LLM–KG integration. Specifically, given the factual knowledge retrieved at each reasoning step, CoG first identifies the corresponding KG schemas and represents these schemas as Python classes, which serve as abstract interfaces to the retrieved facts. It then generates executable code grounded in these classes, with the retrieved facts instantiated as objects of the corresponding classes during execution. This design enables flexible code-based reasoning while avoiding the direct injection of large-scale factual knowledge into prompts.
Experiments on WebQSP, CWQ, and GrailQA demonstrate that CoG outperforms prior state-of-the-art models by up to \textbf{10.5}\%.

\end{abstract}

\section{Introduction}

\begin{figure}[htbp]
\centerline{\includegraphics[scale=0.55]{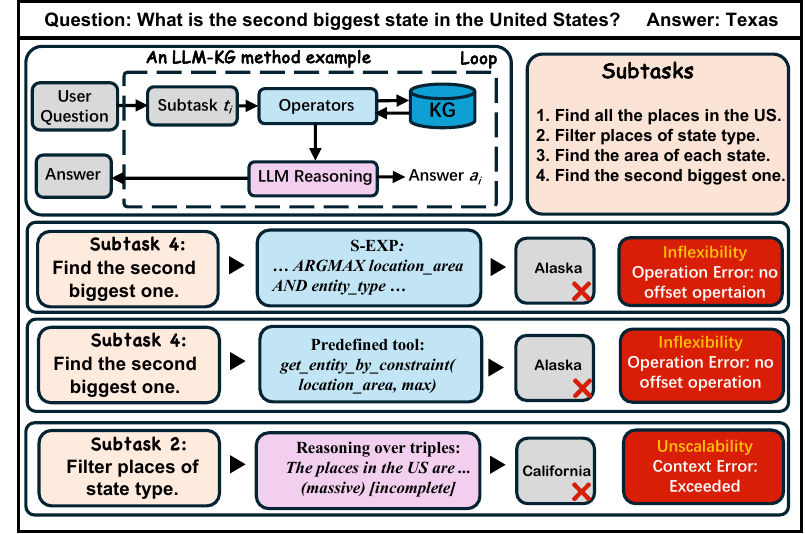}}
\caption{Limitations of the existing LLM-KG methods.}
\label{fig:existed_work}
\end{figure}

Large Language Models (LLMs) \cite{zhao2025llm,cahyawijaya2024fewshotllm} have demonstrated impressive Natural Language Processing (NLP) capabilities, significantly enhancing tasks such as question answering \cite{wang2023qa1,li2023qa2}, text generation \cite{shi2025text2sql}, and textual reasoning \cite{wei2023cot}. 
Despite their success, LLMs remain susceptible to out-of-date knowledge and hallucinations.
Since KGs \cite{ren2007dbpedia,bollacker2008freebase} provide large-scale, structured, and reliable factual knowledge, many studies \cite{sun2024tog,chen2024pog,li-2024-keldar,jiang2024kgagent} have attempted to integrate LLMs with KGs to mitigate these issues.

SPARQL is one of the most widely used query languages for accessing and reasoning over knowledge graphs, offering an expressive interface for specifying graph patterns, joins, constraints, and aggregations over structured facts.
However, directly asking LLMs to generate executable SPARQL queries remains difficult, since it requires accurate schema grounding, compositional graph-pattern construction, variable binding, and strict syntactic correctness \cite{huang-etal-2023-markqa,luo2024chatkbqa}.
Therefore, instead of directly generating SPARQL, prior LLM-KG studies commonly adopt S-expressions as compact intermediate logical forms \cite{gu2021grailqa,shu2022tiara,ye-etal-2022-rng}, or equip LLMs with predefined tools/operators to interact with KGs \cite{sun2024tog,chen2024pog,jiang2024kgagent}.
These alternatives lower the burden of query generation by hiding low-level SPARQL syntax behind a more constrained interaction interface, such as relation traversal, entity retrieval, filtering, and aggregation.

However, such constrained interfaces also lead to two critical bottlenecks.
1) \textbf{Inflexibility}: S-expressions and predefined tools usually restrict reasoning logic to a fixed grammar or operator inventory, including relation traversal, entity retrieval, filtering, and simple aggregation operations such as counting and maximization. However, complex KGQA questions often require more fine-grained and task-specific constraints, such as ranking with offsets, nested filtering, value comparison, or customized aggregation.
As illustrated in Figure~\ref{fig:existed_work}, existing methods fail to answer the question ``What is the second biggest state in the United States?''~\footnote{This question is drawn from the WebQSP dataset.} because they only support the argmax operator.
2) \textbf{Unscalability}: Existing methods usually inject all retrieved KG facts into prompts for LLM reasoning, which limits their performance on large-scale KGs. As reasoning steps increase, retrieved facts grow rapidly, making it increasingly inefficient to inject all facts into prompts.
Meanwhile, context-length constraints severely limit the number of facts that LLMs can actually use for reasoning \cite{luo2024rog}.
As shown in Figure 1, existing methods may exceed the context length limit and fail to include key information relevant to the question.

We observe that KG schemas provide highly abstract representations of large collections of factual knowledge, offering an effective way to guide LLMs to operate over large-scale KG facts.
The key challenge, however, is how to enable LLMs to understand complex schemas and adaptively compose task-specific reasoning operations conditioned on the input question.
Notably, in programming languages such as Python, schemas can be naturally represented as classes, providing a structured and LLM-readable interface \cite{gao2023pal,chen2023pot}.
Reasoning over the corresponding facts can then be implemented through executable functions, allowing LLMs to flexibly compose task-specific operations over large-scale KG facts.
This provides a more flexible and scalable alternative to existing methods that rely on S-expressions or predefined tools.

We instantiate these ideas in Code-on-Graph (CoG), an iterative framework with three stages: (1) Planning, which decomposes a complex question into subtasks; (2) Coding, which retrieves facts, abstracts their schemas into Python class definitions, and generates task-specific executable operations beyond predefined operators; and (3) Executing, which provides the retrieved facts to the code as class instances, runs the generated code in a sandbox environment, and uses execution feedback for correction and retry. In this way, CoG achieves flexibility through dynamic code generation over schema-level abstractions and scalability through compact class-based fact representation.
Our main contributions are as follows:

\begin{itemize}
    \item We propose CoG, an iterative programmatic reasoning framework for LLM-KG that decomposes complex questions into subtasks, generates executable code over schema-level abstractions, and refines the code through execution feedback.
    \item By representing schemas with Python classes and conducting schema-level code generation and execution, CoG enables flexible operations over massive factual knowledge with minimal token usage.
    \item By improving both scalability and flexibility, CoG effectively handles complex queries and outperforms previous state-of-the-art methods by up to 10.5\%.
\end{itemize}

\begin{figure*}[htbp]
\centerline{\includegraphics[scale=0.5]{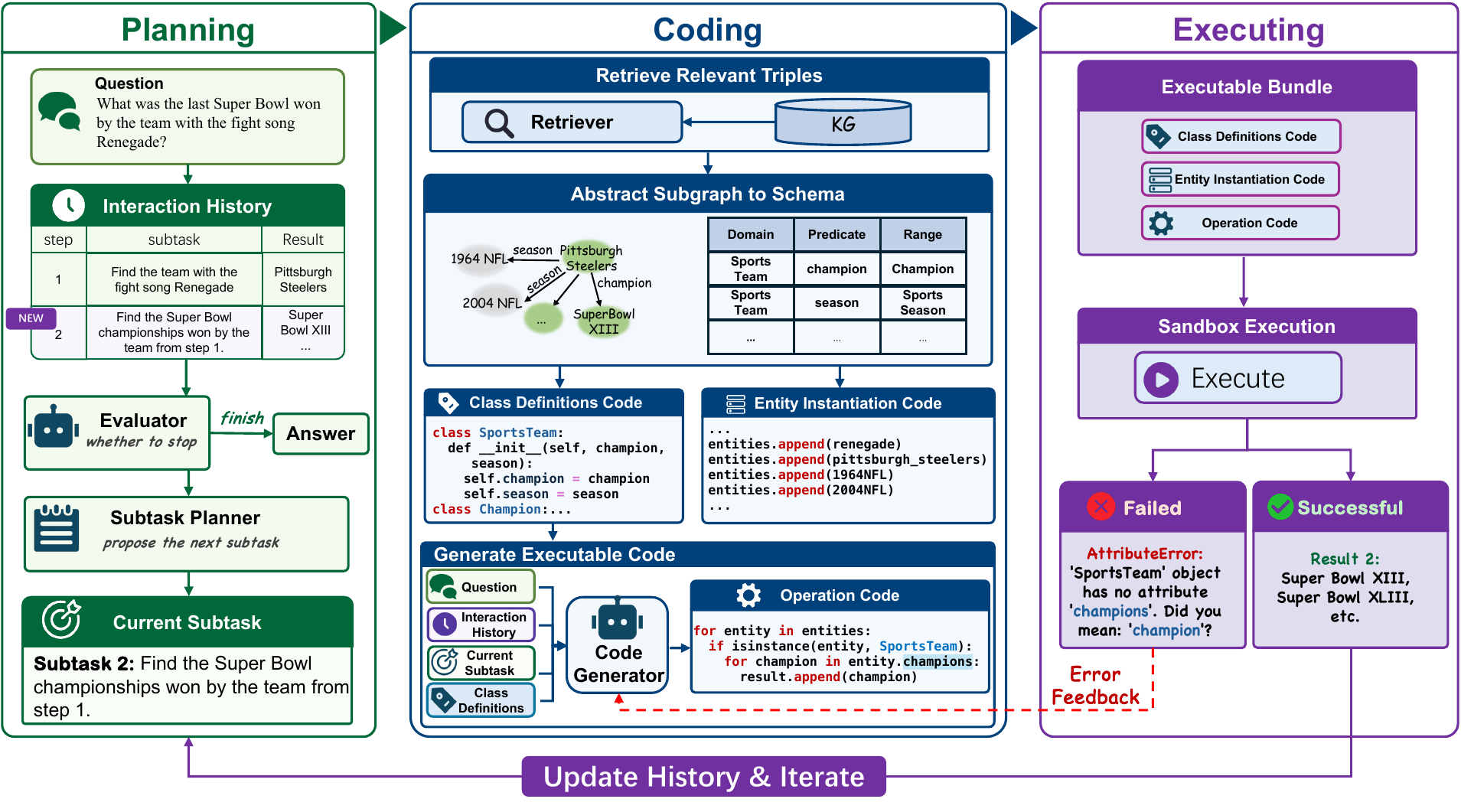}}
\caption{The framework of CoG.}
\label{fig:framework}
\end{figure*}
\label{sec:methodology}

\section{Related Work}
\subsection{LLM-KG Integration} 
Early integration of LLMs with KGs primarily followed the semantic parsing paradigm, where models translate natural language into logical forms like SPARQL or Cypher to query the database \cite{luo2024chatkbqa, li2023kbbinder}. While effective on closed schemas, these methods suffer from brittleness when facing schema inconsistencies or complex multi-hop reasoning requirements. To improve robustness, recent research has shifted toward agentic reasoning frameworks. Representative methods such as Think-on-Graph (ToG) \cite{sun2024tog} and Plan-on-Graph (PoG) \cite{chen2024pog} treat the LLM as an autonomous agent that iteratively explores the graph structure. Similarly, frameworks like StructGPT \cite{jiang2023structgpt}, KG-Agent \cite{jiang2024kgagent}, and RoG \cite{luo2024rog} utilize iterative tool invocation or retrieval-augmented generation (RAG) \cite{he2024gretriever, gao2025d} to navigate external knowledge. 


Compared with conventional semantic parsing, CoG does not aim to generate a single database query in one step. Instead, it treats KG reasoning as an iterative program synthesis problem over retrieved schema abstractions. Compared with agentic tool-use frameworks, CoG does not depend on a small fixed operator inventory to represent all reasoning patterns. Instead, it allows the model to synthesize task-specific executable logic, making intermediate reasoning states explicit and reusable.
\subsection{Program-Aided Reasoning}
This paradigm demonstrates the effectiveness of decoupling reasoning (via program synthesis) from computation (via execution), significantly mitigating hallucination and improving performance on structured-data reasoning, particularly in mathematical, algorithmic, and deep research tasks \cite{gao2023pal, chen2023pot, li2025knowcoder, liu2026towards}. This success has extended to general tool-augmented agents \cite{schick2023toolformer, lu2023chameleon} and neuro-symbolic methods for structured data, such as Binder \cite{cheng2023binder} for tables and ViperGPT \cite{suris2023vipergpt} for visual reasoning. For KG, although code is widely used as a schema representation method in KG construction~\cite{li2024knowcoder, zuo2025knowcoder, guo2024retrieval},
its application in KGQA remains limited. Existing program-aided KG approaches~\cite{chen2024self,10.1145/3774904.3792662} typically rely on generating standard query languages or invoking a static set of predefined function primitives (e.g., rigid hop or filter operators). These fixed operators lack the flexibility to handle the diverse reasoning patterns found in open-domain questions. Inspired by object-oriented programming, our CoG framework advances this field by synthesizing dynamic Python code over abstracted KG objects. This allows the model to "write its own tools" on the fly, offering superior generalization compared to rigid API calls.

\section{Preliminaries}
Following previous work \cite{sun2024tog,chen2024pog},
in this paper, we evaluate CoG on KGQA, a representative and widely studied task for LLM–KG integration that aims to answer natural language questions through reasoning over KGs. 

\subsection{Knowledge Graph}

A KG is a structured representation of factual knowledge, typically modeled as a set of relational triples:
\begin{equation}
\mathcal{G} = \{ (e, r, e') \mid e, e' \in \mathcal{E},\ r \in \mathcal{R} \},
\end{equation}
where $\mathcal{E}$ denotes the set of entities and $\mathcal{R}$ denotes the set of relations. Each triple $(e, r, e')$ represents a factual assertion that relation $r$ holds between a head entity $e$ and a tail entity $e'$.

\subsection{KGQA}
Formally, given a natural language question $q$, a KG $\mathcal{G}$, and a set of topic entities $\mathcal{T}_q \subseteq \mathcal{E}$ mentioned in $q$, the goal of KGQA is to predict a set of answers $\mathcal{A}_q$ that correctly answer the question.

Following common practice in prior work, we assume that topic entities and answer entities are already linked to their corresponding KG identifiers:
\begin{equation}
\mathcal{T}_q, \mathcal{A}_q \subseteq \mathcal{E}.
\end{equation}

This setting isolates the reasoning component of KGQA by bypassing entity linking, allowing the model to focus on semantic understanding and multi-hop inference over the graph.

In many benchmark datasets, each question is associated with a gold reasoning path or an equivalent structured query (e.g., SPARQL), which defines a sequence of relations connecting topic entities to answer entities. However, during inference, such gold paths are not available and must be implicitly or explicitly inferred by the model.

\section{Methodology}

CoG is a programmatic reasoning framework that iteratively decomposes complex questions into subtasks and performs schema-level code-based reasoning over a KG. As illustrated in Figure \ref{fig:framework}, each CoG iteration comprises three core modules: (1) \textbf{Planning}, which decomposes the question into manageable subtasks to progressively reach the answer; (2) \textbf{Coding}, which maps the schemas of retrieved subgraphs into Python classes and generates code based on these classes; and (3) \textbf{Executing}, which executes and refines the code over instantiated classes in a loop to obtain the final results. 



\subsection{Planning}
\label{sec:planning}

The planning module reduces program synthesis difficulty by decomposing a complex question into a sequence of executable subtasks. As illustrated in Figure~\ref{fig:framework}, rather than committing to a rigid, hand-designed decomposition strategy, CoG uses the full interaction history to decide what should be solved next and when reasoning should stop. This makes the reasoning process adaptive to intermediate discoveries and avoids unnecessary exploration.


\paragraph{Subtask Generation.}
To avoid the inflexibility and redundant reasoning paths inherent in static decomposition, the generator $\mathcal{M}_{gen}$ dynamically determines the next subtask by leveraging the full history of prior reasoning. By conditioning on past subtasks and their execution results, the generator avoids unproductive directions and enables non-linear exploration, allowing the current subtask to build upon any relevant previous step rather than strictly following the immediate predecessor. 

Formally, we represent the execution history up to step $t-1$ as $\mathcal{E}_{t-1} = [(T_1, R_1), \dots, (T_{t-1}, R_{t-1})]$, where each pair consists of a subtask $T_i$ and its corresponding execution result $R_i$. The generator then produces the next subtask $T_t$ as:
\begin{equation}
T_t = \mathcal{M}_{gen}(Q, \mathcal{E}_{t-1}).
\end{equation}

Each generated subtask $T_t$ is a structured representation that includes a natural-language description, along with associated topic entities and candidate predicates for retrieval, which facilitate subsequent code generation. Notably, each subtask maintains a parent step reference that specifies the earlier step on which it builds, enabling CoG to capture complex, non-sequential dependencies beyond a strictly linear execution order.

\paragraph{Evaluation.} 
Following the initial generation and execution of the first subtask, the evaluator $\mathcal{M}_{eval}$ determines at each subsequent iteration $t > 1$ whether further exploration is required. By assessing the previously proposed subtasks $\mathcal{T}_{t-1} = [T_1, \dots, T_{t-1}]$ and the latest execution result $R_{t-1}$, the evaluator enables adaptive termination from two complementary perspectives: 
(1) \textbf{Exploration Sufficiency}: whether crucial subtasks decomposed from the question  (e.g., subtasks that satisfy specific constraints such as temporal qualifiers or  superlatives) have been fully addressed; 
(2) \textbf{Answer Plausibility}: whether the most recent result $R_{t-1}$ is type-consistent and semantically aligned with the expected answer form (e.g., entity, numeric value, or date).

Formally, at iteration $t > 1$, the evaluator predicts a binary continuation decision $c_t \in \{\texttt{True}, \texttt{False}\}$ as:
\begin{equation}
c_t = \mathcal{M}_{eval}(Q, \mathcal{T}_{t-1}, R_{t-1}).
\end{equation}
If $c_t = \texttt{True}$, the process triggers the \textit{Generator} to produce $T_t$; otherwise, the planning process terminates and proceeds to summarize the final answer.

\subsection{Coding}
\label{sec:retrieval}
To resolve each subtask $T_t$, the coding module follows three sequential steps: \textbf{subgraph retrieval}, \textbf{schema-class mapping}, and \textbf{code generation}. By compiling large-scale factual knowledge into concise schema-level Python classes, this module enables flexible reasoning through code generation. First, it extracts a relevant subgraph from the KG using the topic entities and candidate predicates associated with the subtask. Next, rather than handling raw triples, the module abstracts the retrieved schemas into Python class definitions. Finally, it generates executable code that instantiates these classes and performs logical operations to resolve the subtask.
\paragraph{Subgraph Retrieval.}
Since reasoning over the entire knowledge graph $\mathcal{G}$ is computationally expensive and prone to noise, this submodule retrieves a compact subgraph $\mathcal{G}_{sub} \subset \mathcal{G}$ based on the topic entities and predicates identified in the subtask $T_t$. Specifically, beginning from the anchor entities, the module performs a bounded expansion with a maximum depth $d$. To control the branching factor, it considers both outgoing and incoming edges for each visited node, retaining only the top-$K$ edges per direction. This neighborhood selection is guided by the semantic similarity between candidate relations and the combined context of the current subtask and its target predicates. An off-the-shelf, pre-trained DistilBERT model \cite{sanh2020distilbert}\footnote{\url{https://huggingface.co/sentence-transformers/msmarco-distilbert-base-tas-b}} is employed to compute these similarity scores.

\paragraph{Schema-to-Class Mapping.}
\label{sec:schema_abstraction}
A key idea in CoG is to transform retrieved KG structures into typed executable abstractions. Directly linearizing triples into text often obscures the graph structure and forces the model to reason over long, noisy contexts. In contrast, CoG maps the retrieved subgraph into a compact set of Python classes, where entity types become classes and relations become typed attributes.

This abstraction preserves the relational structure of the KG while providing an explicit and reusable state representation for subsequent reasoning. As a result, the model can manipulate KG facts through programmatic operations such as traversal, filtering, sorting, aggregation, and constraint checking, instead of relying on fragile natural-language descriptions of the same evidence.



The mapping process is performed in two stages. First, for each predicate $p$ within the retrieved subgraph $\mathcal{G}_{sub}$, the module utilizes schema constraints to infer the class types for the head and tail entities of each triple $(h, p, t)$. Subsequently, each inferred entity type is mapped to the corresponding Python \texttt{class}. For each predicate $p$, a member attribute is added to its corresponding domain class, where the values are instances of the range class. To accommodate multi-valued relations, these attributes are represented as lists. 

Through this mapping, a large volume of raw triples is compressed into a compact set of typed class definitions. This representation preserves the relational structure of the knowledge graph while providing a programmatic interface for the subsequent code generation stage. Consequently, it significantly reduces the token footprint while maintaining high reasoning flexibility for complex queries.

\paragraph{Code Generation.}
Based on the generated Python classes $\mathcal{C}_t$, this step synthesizes the executable logic required to resolve the subtask. Specifically, the LLM is prompted with the original question $Q$, the current subtask specification $T_t$ (comprising topic entities, predicates, and parent step references), and the class definitions $\mathcal{C}_t$. The LLM generates Python code that resolves the subtask primarily through entity traversal, constraint filtering, and data collection. The intermediate results are stored in a predefined container (e.g., a results dictionary). Notably, the code generation process leverages the parent step reference in $T_t$ to access outputs from previous iterations, thereby facilitating non-linear reasoning. 

By operating over the abstract class interface rather than raw triples, the generated code remains concise and less prone to the hallucinations typically associated with long-context factual injection. The final execution of this code yields the subtask result $R_t$, which is then appended to the execution trace $\mathcal{E}_t$.

\subsection{Executing}
\label{sec:program_reasoning}
To obtain the execution result, the module provides an environment to instantiate the Python classes on $\mathcal{G}_{sub}$ and run the code on it. To ensure robustness, the module incorporates a self-correction loop that allows for iterative refinement of the generated logic based on runtime feedback.

\paragraph{Instantiation and Execution.}
The execution is conducted within a restricted Python sandbox. For each subtask $T_t$, the executor primarily concatenates the class definitions $\mathcal{C}_t$, instantiation code, and the operation code. During execution, entities are instantiated into Python objects, and their attributes are populated according to the retrieved triples in $\mathcal{G}_{sub}$. This process produces an in-memory object set $\mathcal{O}_t$ that supports direct attribute traversal, effectively aligning graph-based reasoning with object-oriented execution. 

The executor returns either a valid result $R_t$ or an error feedback signal $\textit{fb}$ (e.g., an exception traceback). Notably, CoG treats an empty result as a failure case, as it typically indicates a logical mismatch between the generated code and the underlying KG constraints.

\paragraph{Self-correction and Early Termination.}
Upon receiving an execution failure or an empty output, the feedback $\textit{fb}$ is fed back to the LLM to trigger code regeneration. This self-correction loop continues for up to $N$ retries. If a valid result $R_t$ is still unobtainable after $N$ attempts, the system performs an early termination and returns the most recent successful candidate answer, thereby preventing unproductive reasoning cycles. Otherwise, a successful pair $(T_t, R_t)$ is appended to the execution trace $\mathcal{E}_t$.

\begin{table*}[ht]
  \centering
  \small  
  \renewcommand{\arraystretch}{1.2} 
  \begin{tabular}{l c c c c c c}
  \toprule
  \multirow{2}{*}{\textbf{Method}} 
  & \multirow{2}{*}{\textbf{WebQSP}} 
  & \multirow{2}{*}{\textbf{CWQ}} 
  & \multicolumn{4}{c}{\textbf{GrailQA}} \\
  \cline{4-7}
  &&& \textbf{Overall} & \textbf{I.I.D.} & \textbf{Compositional} & \textbf{Zero-shot} \\
    \midrule
    \multicolumn{7}{c}{\textit{Fine-tuning Methods}} \\
    \midrule
    UniKGQA & 77.2 & 51.2 & - & - & - & - \\
    TIARA & 75.2 & - & 73.0 & 87.8 & 69.2 & 68.0 \\
    KG-Agent & 83.3 & 72.2 & - & - & - & - \\
    RoG & 85.7 & 62.6 & - & - & - & - \\
    DeCAF &  82.1 & 70.4 & - & - & - & - \\
    Pangu & - &- & 75.4 & 84.4 & 74.6 & 71.6 \\
    FlexKBQA & - &- & 62.8 & 71.3 & 59.1 & 60.6 \\
    
    \midrule
    \multicolumn{7}{c}{\textit{Prompting Methods}} \\
    \midrule
    IO Prompting + Qwen3-Coder-30B-A3B & 63.1 & 42.6 & 33.4 & 33.5 & 23.0 & 37.1 \\
    IO Prompting + DeepSeek-V3.2 & 75.3 & 52.9 & 35.7 & 34.8 & 26.0 & 39.6 \\
    Readi + GPT-4.1-mini & 80.9 & 60.2 & 71.7 & 67.5 & 60.1 & 77.6 \\
    ReKnoS + GPT-4o-mini & 83.8 & 66.8 & 80.5 & - & - & - \\
    SRP + GPT-4.1-mini & 83.6 & 69.0 & 78.8 & 75.8 & 62.6 & 85.8 \\
    PoG + Qwen3-Coder-30B-A3B & 82.9 & 64.1 & 76.8 & 77.9 & 61.1 & 81.9 \\
    PoG + DeepSeek-V3.2 & 83.9 & 72.6 & 71.6 & 68.4 & 56.4 & 78.3 \\
    \midrule
    CoG + Qwen3-Coder-30B-A3B & 76.0 & 60.8 & 82.0 & 82.5 & 77.9 & 83.2 \\
    CoG + GPT-4.1-mini & 85.4 & 72.2 & 89.4 & 89.0 & 83.6 & 91.6 \\
    CoG + DeepSeek-V3.2 & \textbf{88.7} & \textbf{79.1} & \textbf{91.0} & \textbf{90.5} & \textbf{84.2} & \textbf{93.5} \\
    \bottomrule
  \end{tabular}
  \caption{\label{tab:main_results} Main experimental results on WebQSP, CWQ, and GrailQA. The GrailQA results are broken down into Overall, I.I.D., Compositional, and Zero-shot splits.}
  \label{tab:mainresults}
\end{table*}

\section{Experiments}
\subsection{Experimental Setups}
\paragraph{Datasets and Evaluation Metric.}
To evaluate the effectiveness of CoG on complex KG reasoning, we conduct experiments on three widely used multi-hop KGQA datasets: WebQSP \cite{yih2016webqsp}, CWQ \cite{talmor2018cwq}, and GrailQA \cite{gu2021grailqa}, all based on Freebase \cite{bollacker2008freebase}. For GrailQA, we select the same test subset as used in ToG and PoG to ensure consistency across experiments. In line with prior work \cite{sun2024tog, chen2024pog, xu2024gog}, we report exact match accuracy (Hits@1) as our primary metric. Further details on the datasets can be found in Table \ref{tab:datasets}.
\paragraph{Implementation Details.}
Since the baseline methods employ different backbone models, we select Qwen3-Coder-30B-A3B \cite{yang2025qwen3technicalreport}, DeepSeek-V3.2 \cite{deepseekai2025deepseekv32}, and GPT-4.1-mini \cite{openai2025gpt41} for a fairer comparison. For subtask decomposition, we use a temperature of 1.2 to encourage exploration, while the evaluation module employs a temperature of 0 to ensure strict consistency. Code generation and correction are performed with a temperature of 0.3 to yield stable and reliable outputs. The exploration depth is set to 2 to cover complete CVT relations, and the breadth is set to 8 (i.e., up to 8 incoming/outgoing edges per hop).
\paragraph{Baselines.}
We compare CoG against two main categories of methods: fine-tuning methods and prompting methods. For the former, we choose UniKGQA \cite{jiang2023unikgqa}, TIARA \cite{shu2022tiara}, KG-Agent \cite{jiang2024kgagent}, RoG \cite{luo2024rog}, DeCAF \cite{yu2023decaf}, Pangu \cite{gu2023pangu}, and FlexKBQA \cite{li2024flexkbqa} as baselines. For the latter, we choose IO prompting (i.e., the model directly answers questions given some examples) \cite{cahyawijaya2024fewshotllm}, Readi \cite{cheng2024readi}, ReKnoS \cite{wang2025reknos}, SRP \cite{zhu2025srp}, and PoG \cite{chen2024pog} for comparison. Appendix \ref{sec:baselines} provides descriptions of the baselines, and Table \ref{tab:appendix_comparison} compares several other baseline methods.
\subsection{Main Results}
As shown in Table \ref{tab:mainresults}, we compare CoG with baseline methods on three widely used datasets. Overall, CoG consistently achieves the best performance across all three datasets, outperforming both fine-tuning and prompting-based approaches.

\textbf{Schema-level coding enhances the manipulation of large-scale factual knowledge}. Compared to prompting methods like PoG, ReKnoS, and SRP, which rely on reasoning over raw or pruned factual triples, CoG can process a larger volume of evidence programmatically within a compact context by instantiating facts into class structures. This approach significantly enhances fact manipulation capabilities. On the complex multi-hop CWQ dataset, CoG with DeepSeek-V3.2 achieves 79.1\% Hits@1, which represents a 6.5\% absolute improvement over PoG.

\textbf{Flexible schema-level coding enhances generalization ability}. Unlike frameworks such as KG-Agent or Readi, which are restricted by a fixed set of predefined operators, CoG’s schema-level code generation allows the model to generalize to diverse and novel reasoning patterns that static tool-use paradigms cannot accommodate. CoG demonstrates exceptional generalization in the compositional and zero-shot settings of GrailQA. Specifically, CoG with GPT-4.1-mini outperforms Readi by a margin of 23.5\% on the compositional subset. On the zero-shot subset, the performance lead over Readi reaches 14.0\%.

\textbf{Backbone capability significantly impacts performance}. CoG’s performance is strongly correlated with the underlying model's ability to comprehend complex schemas and generate object-oriented code. Owing to the robust workflow design and schema-level coding, CoG achieves competitive results using the Qwen3-Coder-30B-A3B backbone. Meanwhile, the highest performance is reached with powerful backbones like DeepSeek-V3.2. This indicates that CoG’s ceiling scales with the model's programming proficiency; at the same time, its core mechanism of mapping schemas to classes effectively unlocks structured reasoning potential across a wide range of LLM capabilities.
\subsection{Ablation Study}
Experiments are performed on three datasets using DeepSeek-V3.2 as the backbone model. The results are shown in Table \ref{tab:ablation}. w/o Iterative Planning denotes using pre-decomposed subtasks instead of dynamic subtasks for each question.
w/o Error Correction indicates disabling the code correction mechanism and no longer adjusting based on execution feedback from encountered errors. w/o Code Reasoning represents directly injecting triples/facts into prompts instead of using code generation. Since the number of triples can be extremely large—potentially exceeding the context limits of the LLM—we select up to 500 triples most relevant to each subtask based on similarity, aiming to match the token volume used in CoG. w/ JSON Abstraction refers to replacing Python class definitions for triple schema structure with JSON-formatted representations.
The ablation results show that each component contributes to the overall performance of CoG. Specifically, removing iterative planning degrades performance because complex KGQA questions often require adaptive decomposition based on intermediate results. Moreover, disabling error correction significantly degrades performance, showing that executable reasoning benefits from feedback-driven refinement.
Replacing code-based reasoning with direct fact injection also leads to clear drops, suggesting that explicit programmatic manipulation of structured knowledge is more reliable than reasoning over long serialized triples.
Replacing Python classes with JSON format results in a minor performance drop, suggesting that LLMs have a strong understanding of both Python class structures and JSON formats.

\begin{table}[t] 
\centering
\small
\begin{tabular}{@{}lccccccc@{}}
\toprule
\textbf{Method / Variant} & \textbf{WebQSP} & \textbf{CWQ} & \textbf{GrailQA} \\
\midrule
\textbf{CoG} & \textbf{88.7} & \textbf{79.1} & \textbf{91.0} \\
w/o Iterative Planning & 81.7 & 75.1 & 88.9 \\
w/o Error Correction & 85.9 & 67.5 & 85.9 \\
w/o Code Reasoning & 84.9 & 70.5 & 81.6 \\
w/ JSON Abstraction & 88.4 & 76.4 & 87.2 \\
\bottomrule
\end{tabular}
\caption{\label{tab:ablation} Ablation study on three datasets.}
\end{table}

\subsection{Efficiency Study}

We further analyze runtime efficiency to examine whether CoG can scale factual knowledge operations without incurring excessive inference overhead. Beyond standard measures such as the number of LLM calls, token consumption, and time consumption, we introduce \textbf{Token Utility Rate} (\textbf{TUR}) to measure the average number of facts that can be processed per token (including both input and output tokens). Intuitively, TUR captures the amount of useful factual information processed per token, providing a more direct view of whether additional token usage translates into greater factual reasoning capacity. Due to space limitations, we defer the formal definition, counting rules, and calculation protocol of TUR to Appendix~\ref{app:tur}.

As shown in Table~\ref{tab:utilization}, CoG achieves comparable efficiency to PoG in terms of token consumption and runtime, rather than consistently reducing these costs.
Nevertheless, under comparable token and time budgets, CoG operates on substantially more factual knowledge. On CWQ, CoG operates on 20,276 fact units per question, compared with 408 for PoG, while using a similar number of tokens and slightly less runtime, yielding a $40.7\times$ TUR gain. Similar trends appear on WebQSP and GrailQA, where CoG processes 11,213 and 8,148 fact units, achieving $47.2\times$ and $47.6\times$ higher TUR, respectively. CoG also consistently reduces LLM calls across datasets, from 25.1 to 7.0 on CWQ, 13.0 to 3.8 on WebQSP, and 7.5 to 4.2 on GrailQA. These results indicate that CoG can coordinate substantially larger factual knowledge spaces through LLM-generated code, while maintaining efficient inference and avoiding prohibitive context or runtime overhead.

\subsection{Case Study}

\begin{figure}[!t]
\centering
\includegraphics[scale=0.45]{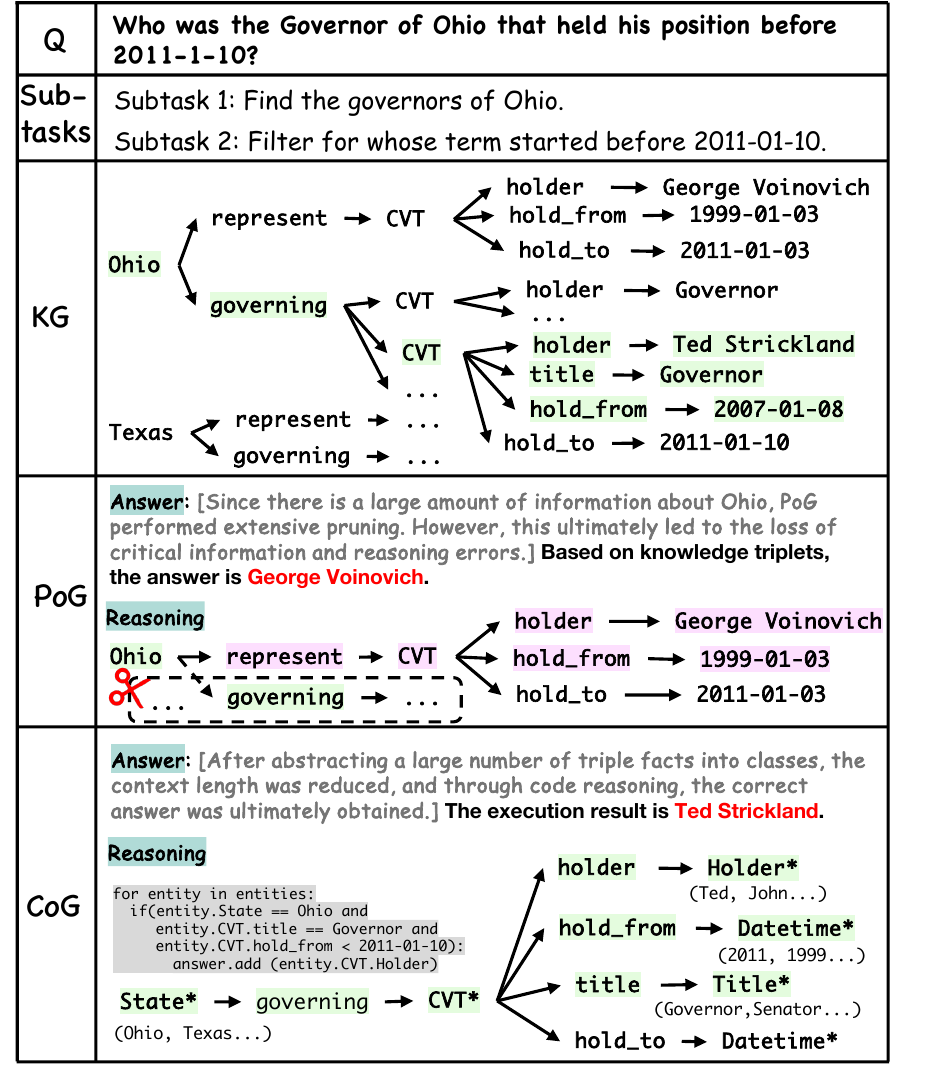}
\caption{An example for comparing PoG and CoG in answering complex questions.}
\label{fig:showcase}
\end{figure}

As shown in Figure \ref{fig:showcase}, we present an example to compare the reasoning processes and results of PoG and CoG. The complex question "Who was the Governor of Ohio that held his position before 2011-01-10?" involves two subtasks: "Find the governors of Ohio" and "Filter for those whose term started before 2011-01-10". The KG contains extensive information about Ohio and its officeholders. When faced with such a large number of facts, PoG reduces the reasoning context by extensively pruning, which ultimately leads to the loss of critical information and results in errors.
In contrast, CoG abstracts entities into a small number of class structures and performs reasoning via code generation and execution.
This design preserves key information and substantially reduces the reasoning context, ultimately leading to the correct answer.

To further demonstrate the flexibility of our method, Appendix \ref{sec:appendix_case_study} presents a comparison with methods relying on predefined tools on questions selected from WebQSP. The results highlight the strong flexibility of our approach while revealing the limitations of predefined tool-based methods.

\section{Conclusion}

In this paper, we introduced a novel LLM-KG framework named Code-on-Graph (CoG), which enhances the reasoning capabilities of LLMs by incorporating KGs. Specifically, we designed CoG to address complex problems through an iterative process of Planning, Coding, and Executing. Extensive experiments demonstrated the effectiveness and efficiency of CoG, which achieved state-of-the-art performance without any fine-tuning.
\section*{Limitations}

CoG can better leverage the programming capabilities of large language models to construct diverse operators for solving complex question-answering tasks. However, its performance is still constrained by the coding strength of the underlying model. When instantiated with a comparatively less powerful coding model such as Qwen3-Coder-30B-A3B, the advantages of CoG are not fully manifested. 

In addition, our experiments only cover a limited subset of LLMs. More state-of-the-art LLMs should be further studied.

\bibliography{main}

\appendix

\section*{Appendix}
\label{sec:appendix}

\section{Dataset Details}
\label{sec:appendix_datasets}

\begin{table}[htbp]
\centering
\small
\begin{tabular}{@{}ccccc@{}}
\toprule
\textbf{Dataset} & \textbf{Answer Format} & \textbf{Train} & \textbf{Test} & \textbf{KB}\\
\midrule

WebQSP & Entity/Number & 3,098 & 1,639 & Freebase\\
CWQ & Entity & 27,734 & 3,531 & Freebase\\
GrailQA & Entity/Number & 44,337 & 1,000 & Freebase\\
\bottomrule
\end{tabular}
\caption{\label{tab:datasets} Dataset Statistics. Due to the large size of the GrailQA test set (6,763 samples) and the associated computational costs, we follow the practice of ToG and PoG by using the same randomly selected subset of 1,000 samples for our evaluation.}
\end{table}

In this work, we employ three complex multi-hop KGQA datasets: WebQSP (WebQuestionsSP) \cite{yih2016webqsp}, ComplexWebQuestions (CWQ) \cite{talmor2018cwq},  and GrailQA \cite{gu2021grailqa}. Table \ref{tab:datasets} summarizes their key statistics.

\textbf{WebQSP} consists of questions from WebQuestions that are answerable via Freebase. The dataset evaluates I.I.D. generalization, containing 34 logical form templates and involving 2,461 entities. It defines 628 relations in total.

\textbf{ComplexWebQuestions (CWQ)} extends WebQSP by introducing four categories of complex questions: conjunction, composition, comparative, and superlative. It is structured over 174 logical form templates, includes 11,422 entities, and covers 845 relations.

\textbf{GrailQA} is a large-scale and diverse KGQA benchmark constructed on Freebase, designed to evaluate model generalization across three distinct subsets based on different settings: I.I.D., compositional, and zero-shot.
\section{Prompts}
\begin{figure*}[htbp]
\centerline{\includegraphics[scale=0.45]{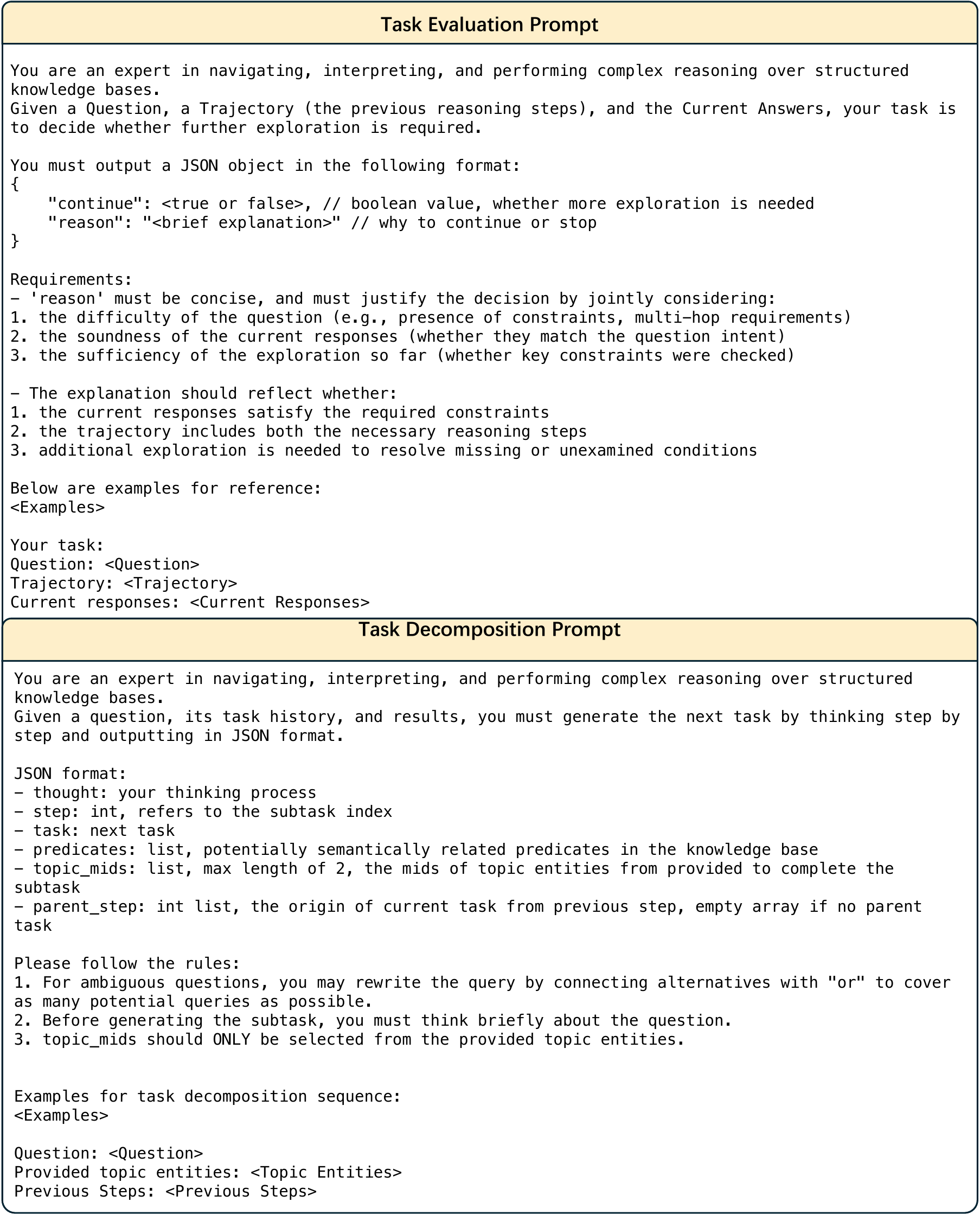}}
\caption{Task decomposition and evaluation prompts.}
\label{fig:task_prompts}
\end{figure*}

\begin{figure*}[htbp]
\centerline{\includegraphics[scale=0.45]{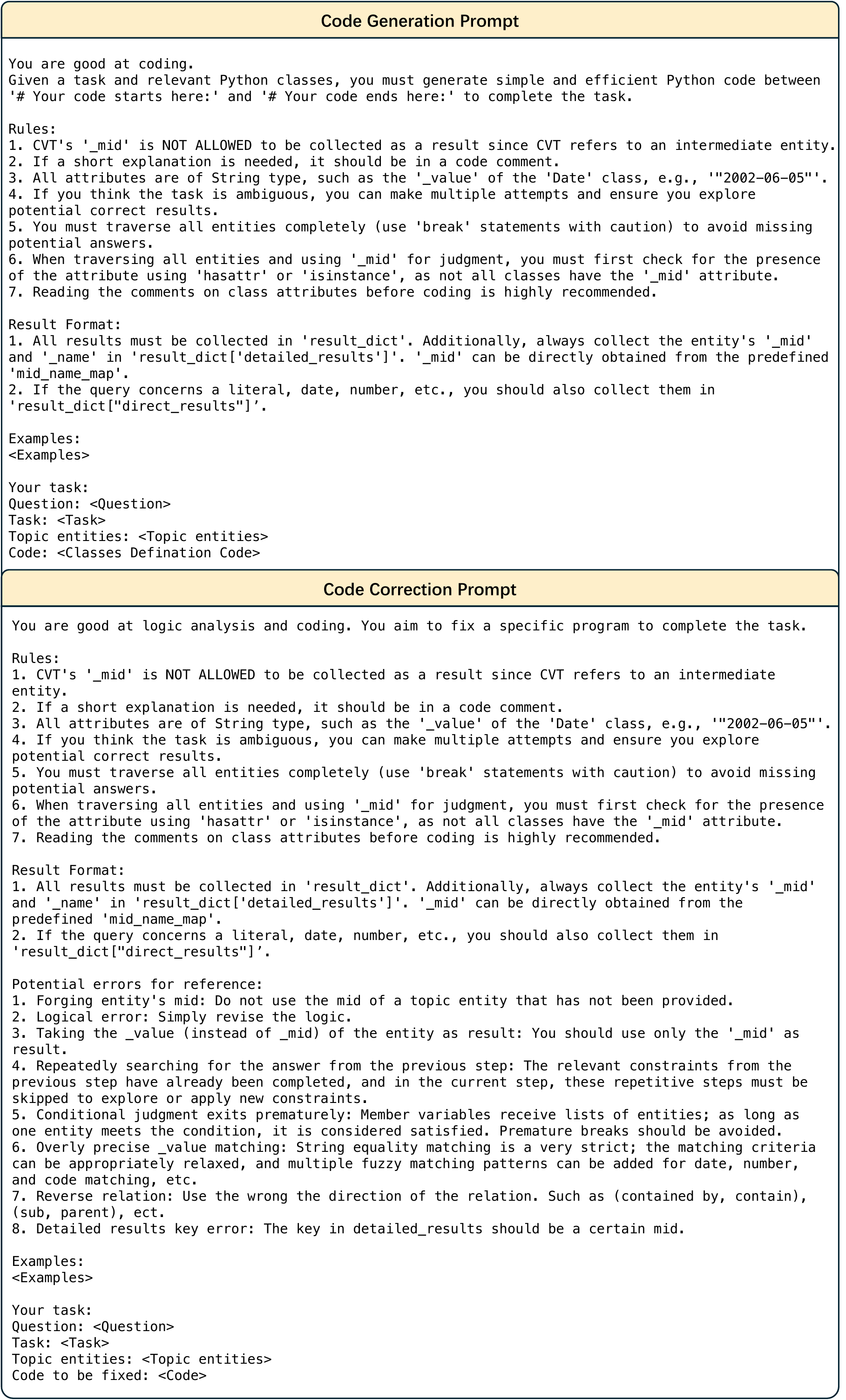}}
\caption{Code generation and correction prompts.}
\label{fig:code_prompts}
\end{figure*}

Figure \ref{fig:task_prompts} presents the prompts related to the planning stage, and Figure \ref{fig:code_prompts} presents the prompts related to the coding and executing stages.

\section{Baseline Descriptions}
In this section, we provide supplementary introductions to baseline methods used for comparison. We categorize them into two groups: Fine-tuning Methods and Prompting Methods.
\label{sec:baselines}
\subsection{Fine-tuning Methods}
\paragraph{UniKGQA} \cite{jiang2023unikgqa} integrates retrieval and reasoning into a single model architecture, utilizing a pre-trained language model for semantic matching between questions and relations, and a matching information propagation module to diffuse relevance signals across the graph. To handle different search scales, the approach introduces abstract subgraphs during retrieval to reduce node complexity. The model is trained via shared pre-training on question-relation matching and stage-specific fine-tuning, effectively bridging the gap between retrieval and reasoning while addressing the challenge of scalable multi-hop QA over large KGs.
\paragraph{TIARA} \cite{shu2022tiara} enhances pre-trained LMs with multi-grained retrieval from knowledge bases. It retrieves relevant entities, exemplary logical forms, and schema items to provide both semantic and structural context for question answering. The approach further employs constrained decoding during generation to ensure syntactic and semantic correctness of logical forms, addressing challenges in grounding and generalization over large KBs.
\paragraph{KG-Agent} \cite{jiang2024kgagent} is an autonomous agent framework that enables smaller LLMs (e.g., LLaMA-7B) to perform complex multi-hop reasoning over KGs. The framework integrates a tuned LLM as a planner, a multifunctional toolbox for KG operations, a KG-based executor, and a knowledge memory module. It employs code-based instruction tuning synthesized from existing KGQA datasets to teach the LLM to autonomously select tools and iteratively reason over KG structures. This approach addresses the limitations of predefined LLM-KG interaction strategies and reduces dependency on large, closed-source LLMs.
\paragraph{RoG} \cite{luo2024rog} synergizes LLMs with KGs to address the issues of hallucination and lack of up-to-date knowledge in LLM reasoning. It introduces a planning-retrieval-reasoning framework where RoG first generates relation paths grounded by KGs as faithful plans, then retrieves corresponding reasoning paths from KGs, and finally conducts interpretable reasoning based on these retrieved paths. By distilling structural knowledge from KGs into the LLM through instruction tuning, RoG enables faithful and interpretable multi-hop reasoning while maintaining the flexibility to integrate with arbitrary LLMs during inference.
\paragraph{DecAF} \cite{yu2023decaf} jointly decodes both logical forms and direct answers within a unified retrieval-augmented sequence-to-sequence model. It linearizes the knowledge base into text passages and employs either sparse (BM25) or dense (DPR) retrieval to fetch relevant context, instead of relying on dedicated entity linking. Using a shared T5-based reader with task-specific prefixes, the model generates logical forms and answers in parallel. The final answer is determined by prioritizing executable logical-form answers when available, otherwise falling back to the directly generated answer. This approach effectively combines the precision of semantic parsing with the robustness of direct generation, while simplifying adaptation across datasets by eliminating the need for specialized entity linking components.
\paragraph{Pangu} \cite{gu2023pangu} leverages LMs primarily as discriminators rather than generators. In this framework, a symbolic agent interacts with the environment to incrementally enumerate and search for valid, executable candidate plans. The neural LM is employed to score and rank these candidate plans based on their plausibility with respect to the input natural language utterance, guiding the search process. This approach shifts the burden of ensuring grammatical correctness and environmental faithfulness from the LM to the agent, effectively addressing key challenges in grounding language to complex, real-world environments.
\paragraph{FlexKBQA}\cite{li2024flexkbqa} first samples diverse, executable programs (e.g., SPARQL queries) from the knowledge base and uses LLMs to convert them into natural language questions, generating a synthetic dataset for training a lightweight model. To bridge the distribution gap between synthetic data and real user questions, it introduces an execution-guided self-training method that iteratively annotates unlabeled user queries. Additionally, the framework incorporates the inherent reasoning capability of LLMs to further enhance robustness and coverage, thereby reducing reliance on extensive manual annotations while maintaining performance across different domains and query languages.
\subsection{Prompting Methods}
\paragraph{IO Prompting} refers to the approach of providing the LLM with several question-answer exemplars and directly prompting it to generate answers. This method relies solely on the internal knowledge of the large model without introducing additional knowledge for assistance. It is prone to generating hallucinations and suffering from outdated knowledge, highlighting the limitations of LLM-based question answering.
\paragraph{Readi} \cite{cheng2024readi} is a framework for efficient and faithful reasoning of LLMs over structured environments like knowledge graphs and tables. It leverages the intrinsic planning ability of LLMs to generate an initial reasoning path, which is then directly instantiated on the environment. By invoking LLMs for path editing only when instantiation fails, Readi reduces iterative LLM calls while using instantiation feedback—such as error locations and candidate relations—to guide corrections. This approach enables LLMs to reason over large-scale structured data with fewer interactions while maintaining grounding fidelity.
\paragraph{ReKnoS} \cite{wang2025reknos} is a novel reasoning framework that enhances the ability of LLMs to perform multi-hop reasoning over knowledge graphs by leveraging the concept of super-relations. Super-relations are groups of semantically related fine-grained relations, which are used to expand the search space and mitigate retrieval failures such as misdirection and depth limitations. The framework iteratively selects and scores candidate super-relations using LLM prompting, enabling simultaneous forward and backward reasoning without exponentially increasing computational cost. By summarizing multiple relational paths into super-relations, ReKnoS effectively addresses the challenge of low retrieval rates in complex knowledge graph question answering tasks.
\paragraph{SRP} \cite{zhu2025srp} integrates iterative planning and reflection by first searching for relevant references to guide the process, then generating and checking initial reasoning paths. It performs knowledge retrieval from KGs and employs a self-reflection mechanism to judge and edit the reasoning paths iteratively until a reliable answer is retrieved. The approach addresses the problem of LLM hallucinations and incomplete reasoning in KGQA by combining structured knowledge retrieval with systematic, feedback-driven path refinement.
\paragraph{PoG} \cite{chen2024pog} is a novel self-correcting adaptive planning paradigm for LLMs augmented with KGs. PoG addresses the limitations of existing KG-augmented LLM methods, such as fixed exploration breadth, irreversible reasoning paths, and susceptibility to forgetting query conditions. The framework iteratively performs task decomposition into sub-objectives, conducts adaptive path exploration on the KG, maintains a dynamic memory of retrieved subgraphs and reasoning states, and employs a reflection mechanism to evaluate and self-correct erroneous paths. By integrating guidance through sub-objectives, continuous memory updates, and reflective backtracking, PoG enables more flexible, reliable, and efficient graph reasoning, effectively enhancing the planning and correction capabilities of LLMs in complex KG question-answering scenarios.

\begin{table*}[t]
\centering
\small
\begin{tabular}{lccccccc}
\toprule
\textbf{Dataset} & \textbf{Method} & \textbf{\# LLM Calls} & \textbf{Total Tokens} & \textbf{Fact Units} & \textbf{Time (s)} & \textbf{TUR} & \textbf{Gain} \\
\midrule
\textbf{CWQ} 
& PoG  & 25.1 & \textbf{17,174.3} & 408    & 103.8 & 0.0238 & $1.0\times$ \\
& Ours & \textbf{7.0}  & 20,928.2 & \textbf{20,276} & \textbf{100.1} & \textbf{0.9688} & $\textbf{40.7}\times$ \\
\midrule
\textbf{WebQSP} 
& PoG  & 13.0 & \textbf{8,401.2}  & 177    & \textbf{52.9} & 0.0211 & $1.0\times$ \\
& Ours & \textbf{3.8}  & 11,270.1 & \textbf{11,213} & 65.1 & \textbf{0.9949} & $\textbf{47.2}\times$ \\
\midrule
\textbf{GrailQA} 
& PoG  & 7.5 & \textbf{4,553.5}  & 70     & \textbf{37.0} & 0.0154 & $1.0\times$ \\
& Ours & \textbf{4.2} & 11,112.5 & \textbf{8,148} & 49.0 & \textbf{0.7332} & $\textbf{47.6}\times$ \\
\bottomrule
\end{tabular}
\caption{Efficiency and Token Utilization Analysis}
\label{tab:utilization}
\end{table*}

\section{Definition of Token Utility Rate}
\label{app:tur}

Token Utility Rate (TUR) measures the amount of question-relevant factual information processed per token consumed by the LLM when answering a specific question. It is designed to evaluate the information-processing efficiency of different methods in knowledge-intensive reasoning tasks.

Given a question $q$, TUR is defined as:
\begin{equation}
\mathrm{TUR}(q) =
\frac{N_{\mathrm{rel}}(q)}
{\mathrm{len}(\mathrm{input}_q) + \mathrm{len}(\mathrm{output}_q)},
\label{eq:tur}
\end{equation}
where $N_{\mathrm{rel}}(q)$ denotes the number of question-relevant factual units accessible to the model during inference, $\mathrm{len}(\mathrm{input}_q)$ denotes the number of tokens in the input prompt, and $\mathrm{len}(\mathrm{output}_q)$ denotes the number of tokens in the generated output.

For a dataset $\mathcal{Q}$, we report the average TUR over all questions:
\begin{equation}
\mathrm{TUR}(\mathcal{Q}) =
\frac{1}{|\mathcal{Q}|}
\sum_{q \in \mathcal{Q}} \mathrm{TUR}(q).
\label{eq:avg_tur}
\end{equation}

\paragraph{Factual Unit.}
A factual unit is defined as the minimal independent knowledge fragment within a knowledge graph that can be accessed by the model during inference. In this work, factual units are instantiated in one of the following forms:
\begin{itemize}
    \item \textbf{Triple format:} $(h, r, t)$, where $h$ is the head entity, $r$ is the relation, and $t$ is the tail entity. For example, $(\texttt{Albert\_Einstein}, \texttt{place\_of\_birth}, \texttt{Ulm})$ indicates that Albert Einstein was born in Ulm.
    \item \textbf{One-hop relation format:} $(h, r)$, where $h$ is the head entity and $r$ is an accessible relation. For example, $(\texttt{Albert\_Einstein}, \texttt{place\_of\_birth})$ indicates that the entity Albert Einstein has a birth-place attribute.
\end{itemize}

\paragraph{Counting Rules.}
We count factual units according to the following rules:
\begin{itemize}
    \item Each complete triple $(h, r, t)$ counts as one factual unit.
    \item Each independent one-hop relation $(h, r)$ counts as one factual unit.
    \item Composite relations are decomposed into multiple independent factual units and counted accordingly.
\end{itemize}

\paragraph{TUR Calculation.}
For our CoG-based method, $N_{\mathrm{rel}}(q)$ is computed based on the number of accessible factual units instantiated from Python classes and passed to the LLM during inference. For PoG, $N_{\mathrm{rel}}(q)$ is calculated by counting all one-hop relations involved in relation pruning and all triples accessed during reasoning. The denominator in Eq.~\ref{eq:tur} is computed as the total number of input and output tokens consumed by the LLM for answering question $q$.

\section{Error Analysis}
\begin{figure}[htbp]
\centerline{\includegraphics[scale=0.35]{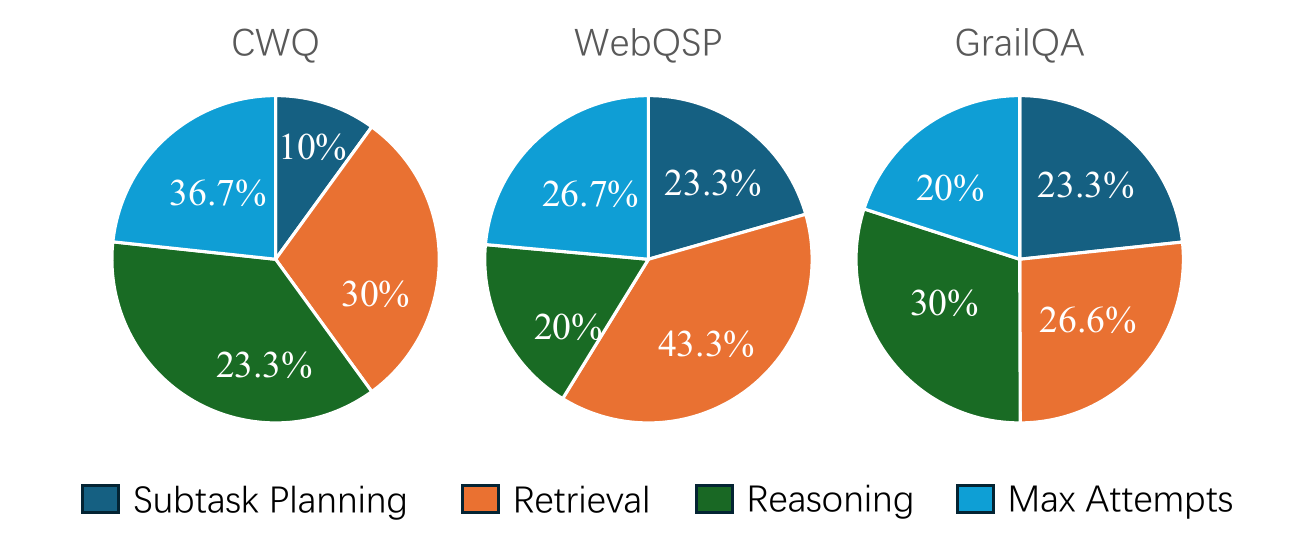}}
\caption{Error distribution across the three datasets.}
\label{fig:error-analysis}
\end{figure}

To systematically analyze our method, we randomly selected erroneous samples from the WebQSP, CWQ, and GrailQA datasets and conducted a manual error analysis. We mainly categorize the errors into four types: Subtask Planning Error, Retrieval Error, Reasoning Error, and Max Attempts Error. Figure \ref{fig:error-analysis} presents the analysis results.

\textbf{Subtask Planning.} This type of error occurs during the subtask decomposition stage. When decomposing complex questions, CoG may deviate from the intent of the original question, or the subtask evaluation module may fail to terminate the reasoning process in time, thereby introducing unnecessary steps that complicate an otherwise straightforward question and ultimately lead to an incorrect answer. For the question ``What sports facility is home to both the Houston Astros and Houston Hotshots?'', CoG correctly decomposes the first three subtasks: (1) ``Find the home sports facility or venue of the Houston Astros'', (2) ``Find the home sports facility or venue of the Houston Hotshots.'', and (3) ``Identify the shared sports facility among the home venues of the Houston Astros and Houston Hotshots.'' At this point, CoG has already identified the correct answer, ``Lakewood Church Central Campus''. However, due to the failure of the evaluation module, it continues to execute additional steps, such as (4) ``Verify whether Lakewood Church Central Campus is a sports facility and confirm that it is the answer to the question asking for the shared home venue of the two teams.'' and (5) ``Find the home sports facility or venue of the Houston Astros for comparison with the venue of the Houston Hotshots.'' These extra steps gradually steer the model away from the correct direction, and after as many as eight reasoning steps, it eventually produces an incorrect answer.

\textbf{Retrieval Error.} This type of error occurs during the subgraph retrieval stage. The retriever fails to recall the correct schema, which leads to failure in the final reasoning process. Such errors usually involve subtasks with implicit semantic meanings.
For example, for the subtask ``Identify the regions the Mississippi River flows through.'', the relation used in the gold query is ``contained by''; that is, if a location contains the Mississippi River, then the river flows through that location. However, CoG fails to recall the relation ``contained by'' for this subtask.

\textbf{Reasoning Error.} This type of error occurs during the code generation and execution stage. Although CoG recalls the correct schema, semantic misunderstanding or the selection of a similar but incorrect relation eventually leads to an incorrect answer.
For example, for the question ``Where does Marta play soccer?'', the gold reasoning chain is pro\_athlete.teams $\rightarrow$ sports\_team\_roster.team $\rightarrow$ [Brazil women's national football team, Tyresö FF]. However, CoG incorrectly reasons through player\_statistics $\rightarrow$ player\_stats\_team $\rightarrow$ [Umeå IK], and finally generates an incorrect answer.

\textbf{Max Attempts Error.} This type of error occurs during the code generation and execution stage. Although CoG recalls the correct schema, the subtask may be challenging, causing the model to become trapped in a local optimum or stuck at a CVT (Compound Value Type) node. This results in repeated correction attempts until the maximum retry limit is reached. Background on CVT: CVT relations represent complex relations and usually require two consecutive hops to move from a semantically meaningful head entity to a semantically meaningful tail entity. The intermediate entity itself has no practical meaning and no readable name. For example, for the question ``Who dated the nominee for the Venice Film Festival Upstream Prize for Best Actress?'', the first subtask, ``Find the nominee for the Venice Film Festival Upstream Prize for Best Actress'', is successfully solved, yielding ``Scarlett Johansson''. However, in the second subtask, ``Find the people who dated Scarlett Johansson, the nominee for the Venice Film Festival Upstream Prize for Best Actress'', the code generated by CoG consistently retrieves meaningless intermediate nodes through the first-hop CVT relation ``celebrity.dated'', without completing the second hop, ``dated.participant'', to obtain the correct answer entities.

\section{Analysis of Code Correction}

\paragraph{Analysis of Different Code Correction Attempts.}During the generation of reasoning code by the LLM, various errors may occur—such as incorrect variable selection, wrong relation extraction, or mistakenly choosing CVT nodes as answers. These issues are corrected through feedback-driven retries. Therefore, we allow CoG a limited number of attempts. We investigate the relationship between the maximum number of attempts per question and Hits@1. 
\begin{figure}[H]
\centerline{\includegraphics[scale=0.35]{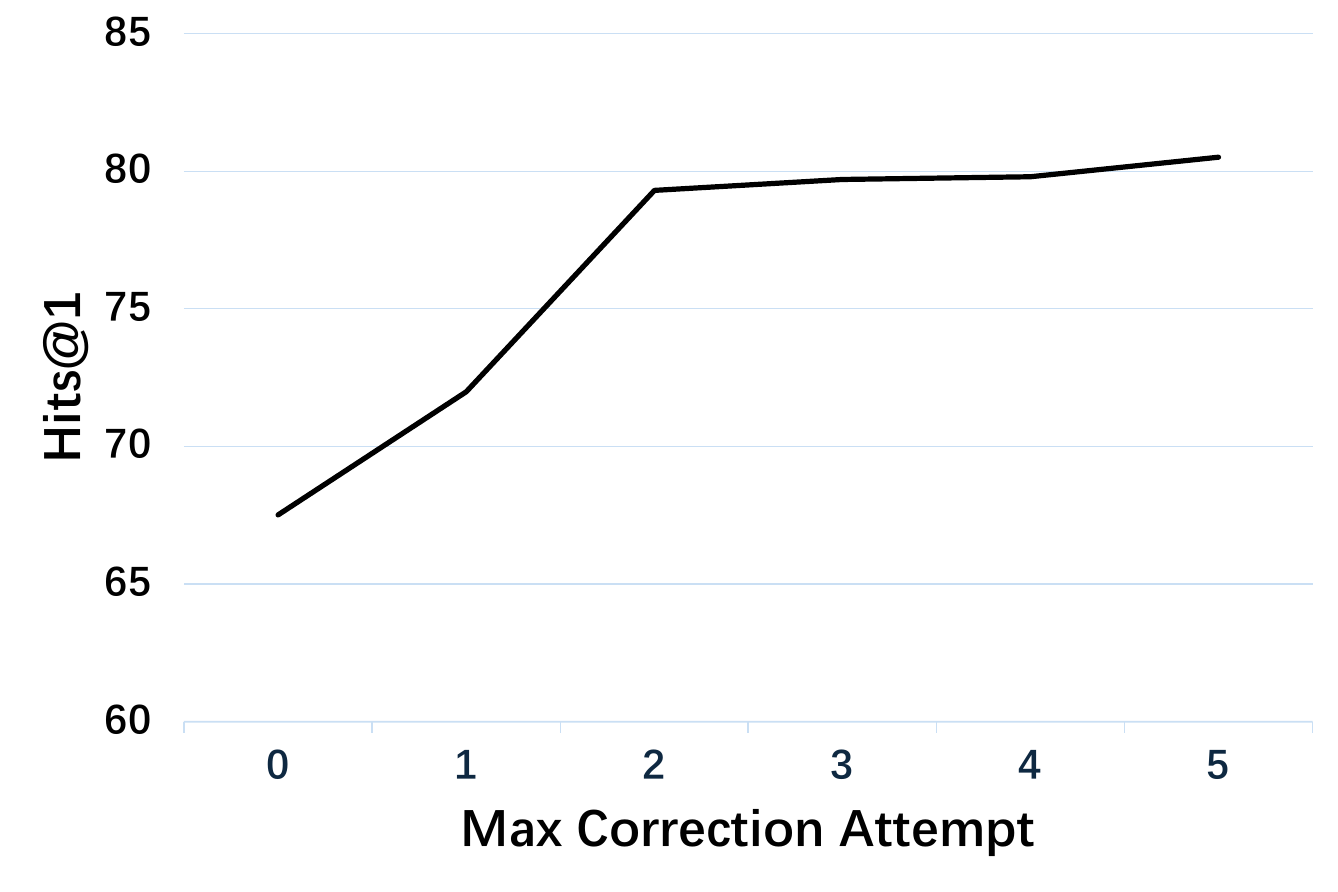}}
\caption{Effect of the Maximum Number of Correction Attempts on CWQ}
\label{fig:attempts}
\end{figure}
As shown in Figure \ref{fig:attempts}, for the more complex CWQ dataset (where each question requires multi-step reasoning), we vary the maximum attempts from 0 to 5. When set to 0, no error feedback or retry is allowed. The results show that increasing the maximum number of attempts consistently improves CoG's performance. However, considering the additional time and token cost incurred by retries, we recommend setting the maximum attempt limit to 3.

\begin{figure}[htbp]
\centerline{\includegraphics[scale=0.35]{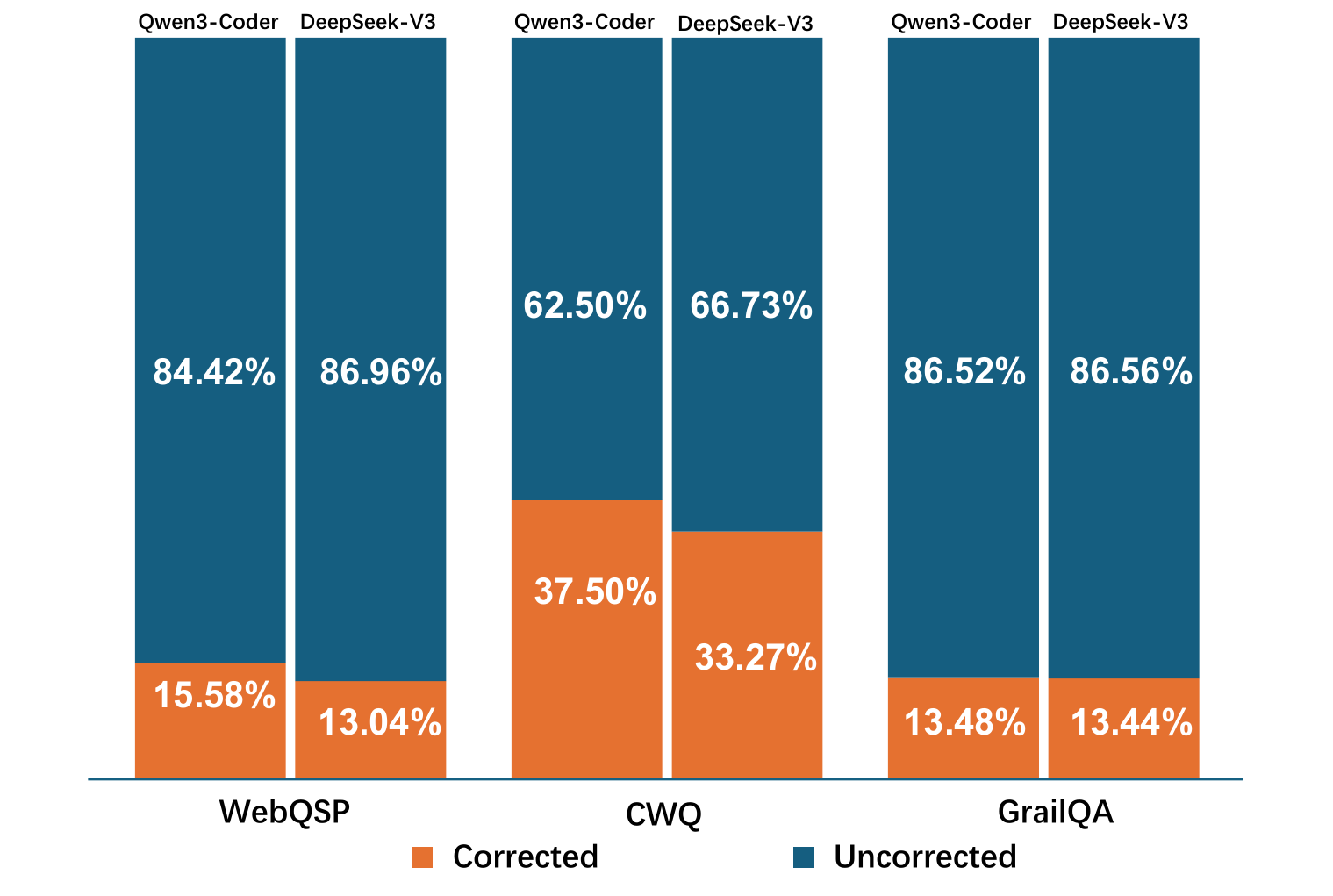}}
\caption{The proportion of cases that involve code correction with Qwen3-Coder-30B-A3B and DeepSeek-V3.2 on three datasets.}
\label{fig:corrected}
\end{figure}

\begin{figure*}[htbp]
\centerline{\includegraphics[scale=0.5]{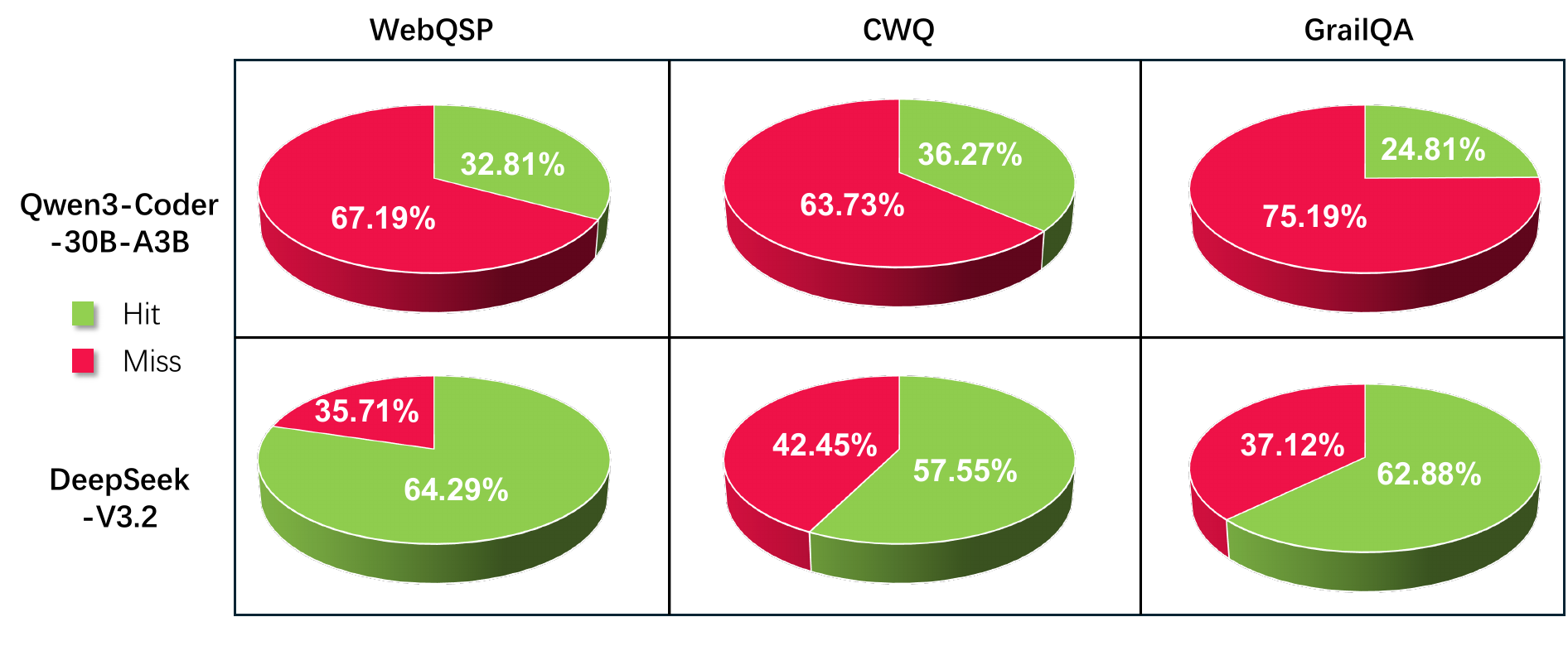}}
\caption{The hit rate of the corrected cases for Qwen3-Coder-30B-A3B and DeepSeek-V3.2}
\label{fig:hit}
\end{figure*}

\paragraph{The analysis of code correction invocation.} Figure \ref{fig:corrected} shows the proportion of code correction invocations by CoG using different models across three datasets. Corrected indicates an invocation, while Uncorrected indicates the opposite. Overall, a significant number of samples invoked the error correction mechanism. CoG's proportion of error correction invocations on WebQSP and GrailQA is lower than that on CWQ, approximately only half of the latter. This is because CWQ has more complex question patterns and greater reasoning difficulty, making it more prone to errors. Examining the performance of different models within each dataset individually reveals that DeepSeek-V3.2, which possesses stronger reasoning and coding capabilities, has a lower proportion of error correction invocations compared to the relatively weaker Qwen3-Coder. This indicates that more powerful models exhibit better stability.

Figure \ref{fig:hit} shows the performance of the samples after the code correction mechanism is applied, specifically whether the final answer is correct. It can be observed that CoG, using both Qwen3-Coder-30B-A3B and DeepSeek-V3.2, successfully corrected many samples, especially the latter, which corrected more than half of the samples. This demonstrates the strong robustness of CoG.

\begin{table}[t]
\centering
\scriptsize
\setlength{\tabcolsep}{3pt}
\resizebox{\columnwidth}{!}{%
\begin{tabular}{lcccccc}
\toprule
\multirow{2}{*}{Method} 
& \multicolumn{2}{c}{WebQSP} 
& \multicolumn{2}{c}{CWQ} 
& \multicolumn{2}{c}{GrailQA} \\
\cmidrule(lr){2-3} \cmidrule(lr){4-5} \cmidrule(lr){6-7}
& Hits@1 & F1 & Hits@1 & F1 & Hits@1 & F1 \\
\midrule
EffiQA  & 82.9 & -- & 69.5 & -- & 78.4 & -- \\
RoG     & 85.7 & 70.8 & 62.6 & 56.2 & -- & -- \\
Ours    & 88.7 & 67.8 & 79.1 & 68.3 & 91.0 & 77.0 \\
\bottomrule
\end{tabular}%
}
\caption{Performance comparison with other methods on WebQSP, CWQ, and GrailQA. We use Hits@1 and F1 score for evaluation. EffiQA \cite{dong2024effiqaefficientquestionansweringstrategic} uses GPT-4, and RoG \cite{luo2024rog} is a fine-tuning method. Our method uses DeepSeek-V3.2.}

\label{tab:appendix_comparison}
\end{table}

\section{Case Study}
\begin{figure}[htbp]
\centerline{\includegraphics[scale=0.5]{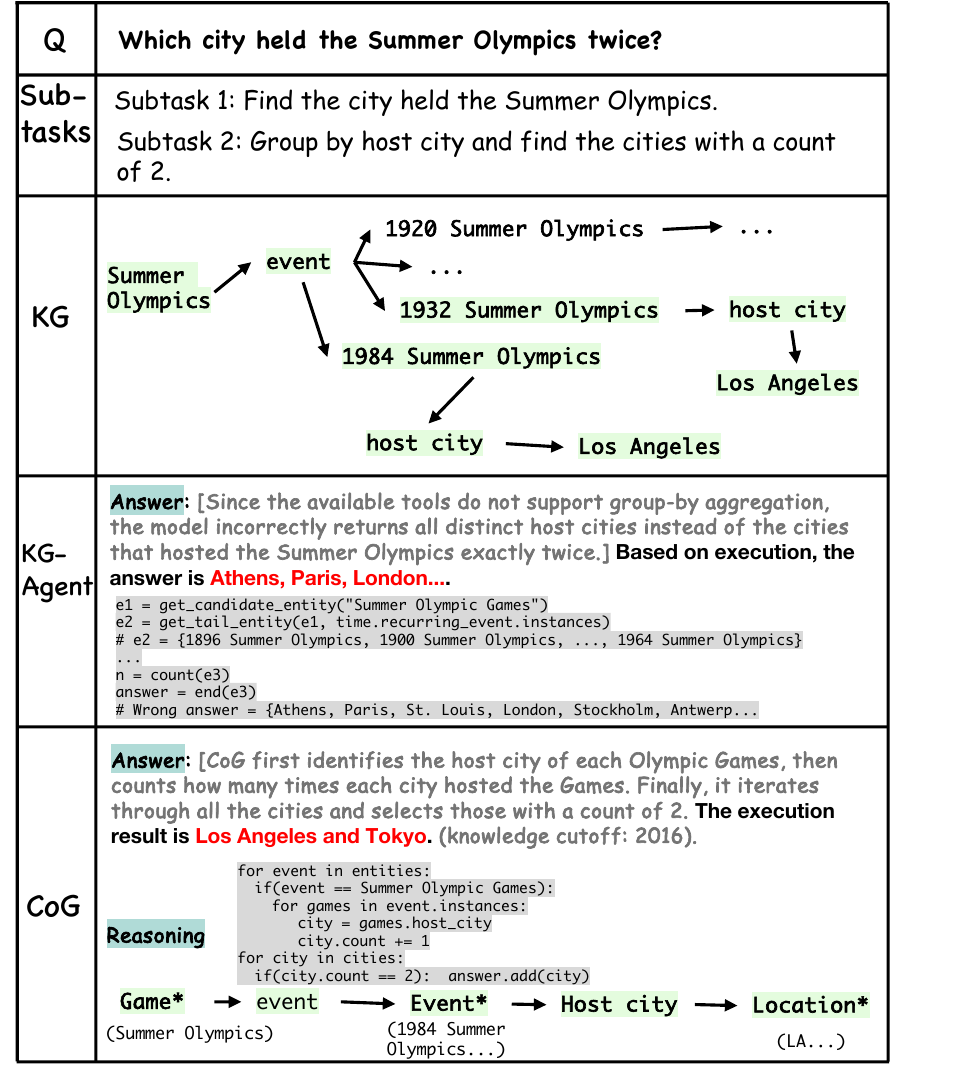}}
\caption{A comparison between CoG and predefined-tool-based methods in answering complex questions. The question is drawn from WebQSP.}
\label{fig:appendix-kg-agent-case}
\end{figure}

\begin{figure*}[htbp]
\centerline{\includegraphics[scale=0.5]{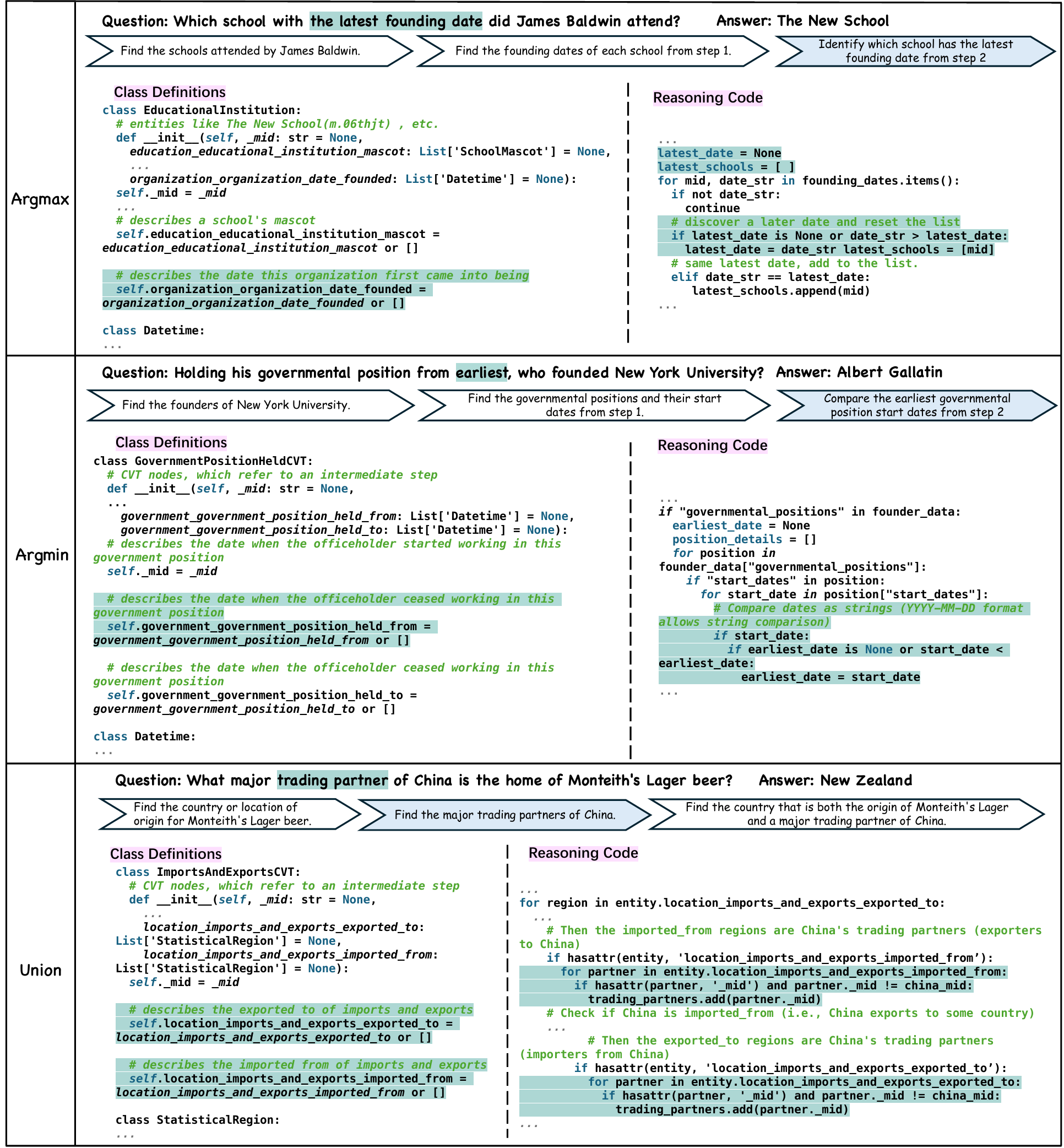}}
\caption{Three representative operators in the CWQ dataset (i.e., Argmax, Argmin, and Union). For each question, we present the scenario corresponding to one of the subtasks. The class definitions of the subtask are shown on the left, while the corresponding operation code is displayed on the right. The parts marked in blue represent the core information relevant to the operation.}
\label{fig:cases}
\end{figure*}

As illustrated in Figure \ref{fig:showcase}, when performing code reasoning, CoG demonstrates rich logical compositions. To demonstrate the advantages of our method, we further compare it with predefined-tool-based methods, as shown in Figure \ref{fig:appendix-kg-agent-case}.
Also, in Figure \ref{fig:cases}, we select three representative scenarios in which CoG generates logical operations, namely Argmax, Argmin, and Union.

For the question “Which school with the latest founding date did James Baldwin attend?”, CoG compares the founding dates of the schools and selects the one with the latest date. For the question “Holding his governmental position from earliest, who founded New York University?”, CoG identifies the founder with the earliest starting date by comparing the dates of holding governmental positions. For the question “What major trading partner of China is the home of Monteith's Lager beer?”, it is necessary to identify countries that either export to China or import from China. CoG accomplishes this operation using OR logic.

\label{sec:appendix_case_study}

\end{document}